\documentclass[runningheads]{llncs}
\usepackage{graphicx}
\usepackage{tikz}
\usepackage{comment}
\usepackage{amsmath,amssymb}
\usepackage{color}
\usepackage[accsupp]{axessibility} 
\usepackage{graphicx}
\usepackage{booktabs}
\usepackage{multirow}
\usepackage{float}
\usepackage{epsfig}
\usepackage{bm}
\usepackage{color}
\usepackage{ulem}
\usepackage[ruled,linesnumbered]{algorithm2e}
\usepackage{hyperref}

\newenvironment{packed_itemize}{
	\begin{itemize}
		\setlength{\itemsep}{1pt}
		\setlength{\parskip}{0pt}
		\setlength{\parsep}{0pt}
	}{\end{itemize}}

\usepackage{xspace}
\makeatletter
\DeclareRobustCommand\onedot{\futurelet\@let@token\@onedot}
\def\@onedot{\ifx\@let@token.\else.\null\fi\xspace}

\makeatother

\usepackage{amsmath}

\usepackage[capitalize]{cleveref}
\usepackage[symbol]{footmisc}

\begin{document}
\pagestyle{headings}
\mainmatter
\def\ECCVSubNumber{4534}  % Insert your submission number here

\title{
Quasi-Balanced Self-Training on Noise-Aware Synthesis of Object Point Clouds for Closing Domain Gap
}

\begin{comment}
\titlerunning{ECCV-22 submission ID \ECCVSubNumber} 
\authorrunning{ECCV-22 submission ID \ECCVSubNumber} 
\author{Anonymous ECCV submission}
\institute{Paper ID \ECCVSubNumber}
\end{comment}

% \maketitle

% \titlerunning{ECCV-22 submission ID \ECCVSubNumber} 
% \authorrunning{ECCV-22 submission ID \ECCVSubNumber} 
% \author{Anonymous ECCV submission}
% \institute{Paper ID \ECCVSubNumber}

%******************

%\begin{comment}
\titlerunning{Quasi-Balanced Self-Training on Noise-Aware Synthesis ...}

\author{
{Yongwei Chen \inst{1,2}}\protect\footnotemark[1] \and
Zihao Wang \inst{1}\protect\footnotemark[1] \and
Longkun Zou \inst{1} \and
Ke Chen \inst{1,3}\protect\footnotemark[2] \and
Kui Jia\inst{1,3}\protect\footnotemark[2]
}
\authorrunning{Y. Chen et al.}
\institute{South China University of Technology \and
DexForce Co. Ltd. \and Peng Cheng Laboratory \\
\email{\{eecyw,eezihaowang,eelongkunzou\}@mail.scut.edu.cn}, \email{\{chenk,kuijia\}@scut.edu.cn}}
% \end{comment}
%******************
\maketitle
\renewcommand{\thefootnote}{\fnsymbol{footnote}}
\footnotetext[1]{Equal contribution}
\footnotetext[2]{Corresponding authors}
    \begin{abstract}
Semantic analyses of object point clouds are largely driven by releasing of benchmarking datasets, including synthetic ones whose instances are sampled from object CAD models. However, learning from synthetic data may not generalize to practical scenarios, where point clouds are typically incomplete, non-uniformly distributed, and noisy. Such a challenge of Simulation-to-Reality (Sim2Real) domain gap could be mitigated via learning algorithms of domain adaptation; however, we argue that generation of synthetic point clouds via more physically realistic rendering is a powerful alternative, as systematic non-uniform noise patterns can be captured. To this end, we propose an integrated scheme consisting of physically realistic synthesis of object point clouds via rendering stereo images via projection of speckle patterns onto CAD models and a novel quasi-balanced self-training designed for more balanced data distribution by sparsity-driven selection of pseudo labeled samples for long tailed classes. Experiment results can verify the effectiveness of our method as well as both of its modules for unsupervised domain adaptation on point cloud classification, achieving the state-of-the-art performance. 
{Source codes and the SpeckleNet synthetic dataset are available at \href{https://github.com/Gorilla-Lab-SCUT/QS3}{https://github.com/Gorilla-Lab-SCUT/QS3}.}

\end{abstract}
    %%%%%%Introduction
\section{Introduction}
\label{sec:intro}

As raw observations of 3D sensors, object point clouds are popularly used for a variety of 3D semantic analysis tasks, including shape classification \cite{PointNet,Pointnet++,DGCNN,SimpleView,RSCNN}, part segmentation of object surface \cite{PointNet,Pointnet++,li2018pointcnn}, estimation of object poses in indoor scenes \cite{Chen2021FSNetFS,Gao20206DOP,Lin_2021_ICCV}, and 3D detection in autonomous driving \cite{li2016vehicle,yang2018pixor,Deng_2022_CVPR}.
The current progress along this line has largely been driven by the publicly released synthetic benchmarks (\textit{e.g.}, the ModelNet \cite{ModelNet} and the ShapeNet \cite{shapenet}).
Object point cloud instances in those datasets are sampled from object CAD models, which are ideally noise-free and uniformly distributed on the complete object surface. 
Very high accuracies (\textit{e.g.}, $96.2\%$ on classification accuracy of the DGCNN \cite{DGCNN} with the ModelNet10 dataset {\cite{PointDAN}}) can be gained by a large number of recent point classifiers \cite{PointNet,DGCNN,Pointnet++,RSCNN,SimpleView,lin2021object}.
However, point clouds collected in real-world applications are typically incomplete, non-uniformly distributed and noisy, due to unavoidable sensor noises and interaction with contextual objects. 
In this way, classification of real-world object point clouds can be difficult, as pointed out in recent works \cite{scanobjectnn,xu2020classification} (\textit{e.g.}, $78.4\%$ on classification accuracy by the DGCNN with the realistic ScanNet10 \cite{PointDAN}).

\begin{figure}[t]
  \centering
   \includegraphics[width= 1.0\linewidth]{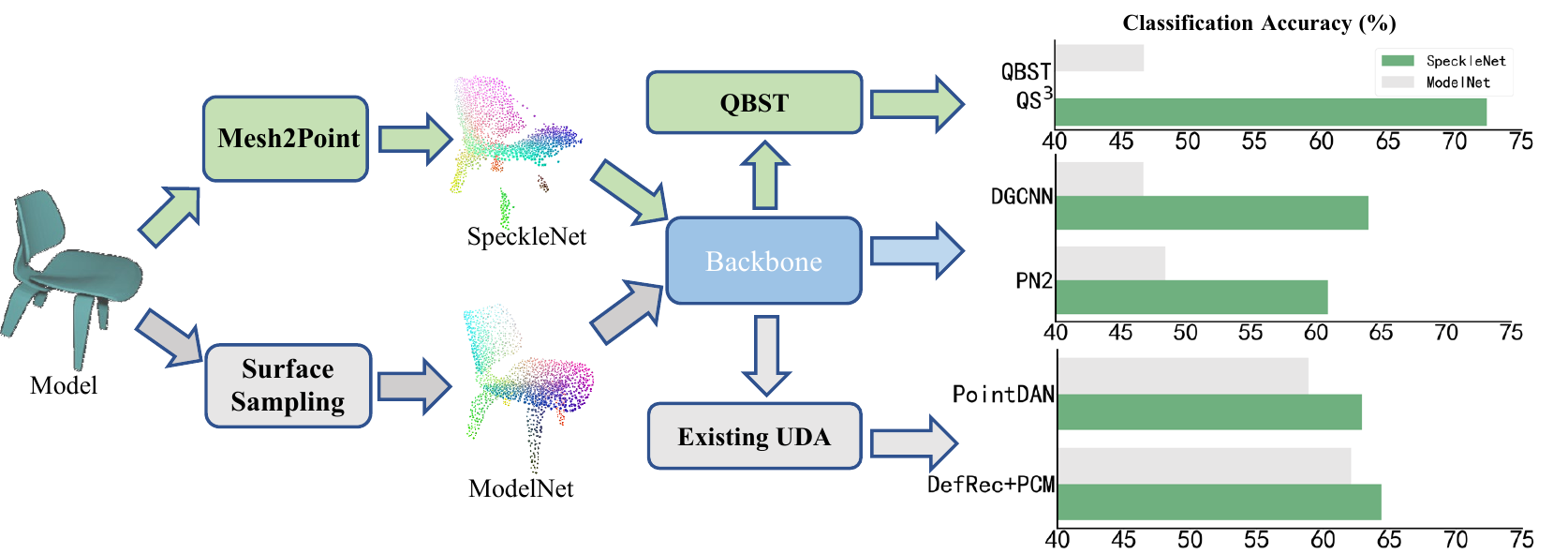}
    \caption{We propose an integrated scheme of Quasi-balanced Self-training on Speckle-projected Synthesis (QS$^3$) to cope with shape and density shift between synthetic and real point clouds. Given identical CAD models, we generate a point cloud SpeckleNet dataset simulating realistic noises in stereo imaging and matching; while point clouds in existing ModelNet dataset \cite{ModelNet} are sampled from object surface of those models. Moreover, we design a novel quasi-balanced self-training (QBST) strategy to further boost the UDA performance. In comparison with two representative UDA methods (DefRec+PCM \cite{DefRecPCM}, PointDAN \cite{PointDAN}) and two representative point cloud classification networks (PointNet++ \cite{Pointnet++}, DGCNN \cite{DGCNN}), our integrated QS$^3$  can perform consistently better, when evaluating on real world data -- an adapted DepthScanNet dataset.}
   \label{fig:intro}
\end{figure}

Classification of object point clouds is made more challenging when considering Sim2Real domain discrepancy of semantic patterns (\textit{i.e.}, training point classifiers on synthetic data and testing on real data), which has been investigated in recent works \cite{DANN,PointDAN,DefRecPCM,GAST,zhang2022masked}.
Such a problem can be formulated into a practical unsupervised domain adaptation (UDA) problem, where supervision signals are only available for source synthetic data. 
Existing methods either adapt effective UDA algorithms on 2D images to point clouds directly as one application \cite{DANN} or capture domain-invariant geometry patterns in self-supervised learning style \cite{GAST,DefRecPCM}. 
Although these methods have gained remarkable success on alleviating the suffering of Sim2Real domain gap in the style of point distribution shift, our paper attempts to address the challenge in a new perspective -- an integrated scheme of physically realistic
synthesis of object point clouds for domain-adapted feature learning.
Our motivation is based on the following observation: point distribution shift can be composed of two types of point deformation from ideal synthetic point clouds: 1) shape variations along the normal directions of surface;
and 2) density variations along the tangent plane of object surface.

On one hand, in view of data-driven nature of deep models, we argue that generation of synthetic point clouds to physically simulate realistic noises is a powerful alternative for coping with the former shape variations, which can intuitively be interpreted as input transformation to align geometric patterns of point clouds. 
Recently, in addition to the straightforward simulation by adding random noises, data augmentation on synthetic point clouds can only mimic partiality and non-uniformness of real point clouds by deformation on a (sub)set of points, whose output is termed as augmented point clouds in this paper.
However, 
only a few works \cite {blensor,planche2017depthsynth,planche2021physics} focus on simulating realistic non-uniform systematic sensor noises in photorealistic rendering and depth image generation, whose output is utilized to convert point clouds.
In view of this, we introduce a Mesh-to-Point (Mesh2Point) pipeline to generate synthetic point clouds based on a photorealistic rendering of a stereo camera. 
{Specifically speaking, by using physically based rendering to simulate 
the workflow of a real-world depth scanner, our Mesh2Point can approach non-uniform systematic noises of depth sensors affected by projection patterns and the method of depth computation.}

On the other hand, physically realistic synthesis of object point clouds via a virtual depth sensor cannot ensure that similar point densities to those of realistic target data, which encourages us to use UDA techniques to further bridge Sim2Real domain gap.
Inspired by positive effects of balanced data distribution to suppress negative transfer \cite{crest,kim2020distribution,Tang_2020_CVPR}, we design a novel quasi-balanced self-training (QBST) with updating the sample selector with pseudo labeled target samples only, which can effectively model geometric patterns of real point clouds in target domain.
Note that, the class balanced sampling strategy to ease the long-tailed data distribution has been widely adopted in different fields, but we argue that it remains non-trivial in the context of UDA.
Its main challenge lies in learning to construct a class-balanced self-training set with diverse samples via selection and pseudo annotation on unlabeled target samples, whose data distribution is unknown.
Our QBST method is simple yet effective, owing to 1) sparsity-sensitive weight sampling filtered by a confidence threshold; and 2) self-training only with pseudo labeled target samples, thus resulting in robust performance against long tailed distributions. 
Experiments on a raw benchmark -- DepthScanNet10 of the ScanNet, without any pre-processing on object samples except resizing, can confirm that our integrated scheme of Quasi-balanced Self-training on Speckle-projected Synthesis (QS$^3$) can significantly improve performance on the challenging Sim2Real UDA task, as shown in Fig. \ref{fig:intro}.

Main contributions of this paper are as follows.
\begin{packed_itemize}
\item We propose a unique QS$^3$ scheme of integrating a  physically realistic synthesis of object point clouds and a UDA point classifier, to jointly cope with shape and density shift between synthetic and real point clouds.
\item A new synthetic dataset  -- the SpeckleNet is constructed via the Mesh2Point generation, which can be readily scaled given sufficient CAD models.
\item A novel quasi-balanced self-training method is proposed to inhibit negative transfer, 
owing to balanced data distribution via sparsity-driven selection of pseudo-labeled target samples. 
\item Experiment results of the proposed QS$^3$ can achieve the state-of-the-art performance on the challenging Sim2Real domain adaptation task.
\end{packed_itemize}

    %%%%%%Related work
\section{Related Work} \label{sec:relatedwork}

\noindent\textbf{UDA on Point Cloud Classification --}
Recently, a few works \cite{PointDAN,DefRecPCM,GAST} propose the problem of UDA on an irregular point-based representation, which inherits the challenge of semantic gap as other UDA problems on images and also has its specific challenge of domain-agnostic feature encoding from local geometries of point clouds. 
Qin \textit{et al.} \cite{PointDAN} explore the first attempt of UDA on point cloud classification by explicitly aligning local features across domains. 
Achituve \textit{et al.} \cite{DefRecPCM} propose to incorporate domain-insensitive local geometric patterns, in a self-supervised reconstruction from partially distorted point clouds, into a global representation, with a simple yet effective data augmentation of point cloud mixup to inherently alleviate imbalanced data distribution. 
Zou \textit{et al.} \cite{GAST} combine self-paced self-training for preserving intrinsic target discrimination and self-supervised learning for domain invariant geometric features, which can capture both global and local geometric patterns invariant across domains.
Different from existing methods concerning superior cross-domain generalization via feature alignment, in our paper, we propose an integrated scheme consisting of the synthesis of realistic point clouds and a quasi-balanced self-training, which can be interpreted as alignment in both input and feature space, to reduce negative effects of domain gap.  

\noindent\textbf{Generation of Synthetic Data --}
Different from benchmarking image classifiers on real images, recent progress of geometric deep learning on point clouds has been largely driven by synthetic datasets \cite{shapenet,ModelNet,blenderproc,heindl2021blendtorch,scenenet,replica,ICL-NUIM,pbrprinceton}, in view of difficulties in acquiring and annotating real-world 3D data. 
Recent works \cite{blenderproc,pbrprinceton,Interiornet} mainly focus on utilizing Physically Based Rendering (PBR) for the synthesis of photorealistic RGB images, while depth images as a byproduct are typically noise-free or simply perturbed by Gaussian noises, which are evidently different from non-uniform systematic noises in practice.
Such an observation encourages a number of works to explore physical simulation of 3D sensors to approach realistic noises. 
Early attempts \cite{ICL-NUIM,kinectbarron,kinectbohg,nguyen2012modeling} in a theoretical style fail to cover all kinds of realistic noises.
Recently, a number of works including ours prefer the Physically Based Rendering owing to its capability of replicating realistic systematic noises.
Specifically, based on PBR, Blensor \cite{blensor} as well as \cite{lidarfang,lidarsim,lidarTallavajhula,reitmann2021blainder} are able to simulate time-of-flight based 3D sensors while other works \cite{blazer,landau2015simulating,planche2017depthsynth,planche2021physics}
mainly focus on simulation of structured light based ones. 
The most relevant works to ours are DepthSynth \cite{planche2017depthsynth} and DDS \cite{planche2021physics}, as all of these methods employ speckle pattern projection in PBR
for realistic depth acquisition during simulation. 
However, both of DepthSynth \cite{planche2017depthsynth} and DDS \cite{planche2021physics} concern about realistic simulation of depth sensors only, while the goal of our synthesis in the context of UDA is to mitigate domain gap, coupled with a new quasi-balanced self-training in a unified QS$^3$ scheme.

\noindent\textbf{Self-training --}
Self-training utilizes pseudo labels generated from predictions of a model learned on labeled data, as supervision signals for unlabeled data, which is widely  used in semi-supervised learning \cite{pseudo_label,FixMatchSSL,arazo2020pseudolabeling} and UDA \cite{triTraining,dwt_mec}. 
Sohn \textit{et al.} \cite{FixMatchSSL} use pseudo labels predicted from a weakly-augmented input image as supervision of its strongly-augmented counterpart; for semantic segmentation on 2D images, while Zou \textit{et al.} \cite{cbst} propose a self-paced learning based self-training framework (SPST), which is formulated as a self-paced learning with latent variable objective optimization \cite{SelfPaceL}.
However, these methods directly choose pseudo-labels with high prediction confidence, which will result in model bias towards easy classes and thus ruin the transforming performance for the hard classes. 
To solve this problem, a class-balanced self-training (CBST) scheme is proposed in \cite{cbst} for semantic segmentation, which shows comparable domain adaptation performance to the best adversarial training based methods. 
The method in \cite{smpcst} proposes a self-motivated pyramid curriculum domain adaptation method using self-training. 
More recently, CRST \cite{crst} further integrates a variety of confidence regularizers to CBST \cite{cbst}, producing better domain adaption results. 
Our QBST method shares similar spirit with CBST \cite{cbst} to neglect negative transfer caused by imbalanced data distribution, but can encourage a more balanced self-training dataset of high confident samples, regardless of source data distribution.

    \section{Methodology}
For unsupervised domain adaptation on point cloud classification, 
given a labeled source domain $\mathcal{S} = \{\mathcal{P}_i^s\}_{i=1}^{n_s}$ with their corresponding class labels $\{y^s\}_{i=1}^{n_s} \in \mathcal{Y}$ and an unlabeled target domain $\mathcal{T}= \{\mathcal{P}_i^t\}_{i=1}^{n_t}$, where $n_s$ and $n_t$ denote the size of samples in source and target domains respectively and point cloud $\mathcal{P}\in \mathcal{X}$ consists of a set of 3d coordinates covering object surface, the semantic label space $\mathcal{Y}$ is shared between both domains.
The objective of UDA on point sets is to learn a domain-adapted mapping function $\Phi: \mathcal{X} \rightarrow \mathcal{Y}$ that classifies any testing sample $\mathcal{P}$ from target domain $\mathcal{T}$ correctly into one of $K = |\mathcal{Y}|$ object categories.
In the context of deep learning, the mapping function $\Phi$ can be decomposed into a cascade of a feature encoder $\Phi_\text{fea}: \mathcal{X} \rightarrow \mathcal{F}$ for any input $\mathcal{P}$ and a classifier $\Phi_\text{cls}: \mathcal{F}\rightarrow \mathcal{Y}$ as follows:
	$\Phi(\mathcal{P}) = \Phi_\text{cls}(\bm{F}) \circ \Phi_\text{fea}(\mathcal{P})$,
where the feature output $\bm{F} \in \mathcal{F}$ of $\Phi_\text{fea}(\mathcal{P})$ and $\mathcal{F}$ denotes feature space. 

Existing UDA classifiers on point sets \cite{PointDAN,DefRecPCM,GAST} concern on reducing domain discrepancy of feature encoding $\Phi_\text{fea}$, while our paper starts from the origin of Sim2Real domain gap, which is caused by point distribution shift of the shape representations of CAD models.
In other words, given an identical CAD model, the procedure of generating synthetic point clouds determines Sim2Real domain gap.
As a result, the problem of this paper can be reformulated as the following: given a set of labeled CAD models $\mathcal{M}=\{M_i, y_i\}_{i=1}^{n_s}$ as source domain $\mathcal{S}$ and an unlabeled set of real point clouds $\{\mathcal{P}_i^t\}_{i=1}^{n_t}$ as target domain $\mathcal{T}$, the point cloud of source domain $\{\mathcal{P}_i^s\}$ are generated from each CAD model $M$. The goal is to learn a feature mapping from point clouds $\mathcal{P}\in \mathcal{X}$ to the labels $y \in \mathcal{Y}$. 
Most of existing synthetic point clouds are directly sampled from CAD models' surface and optionally with moderate pre-defined noises, while the remaining ones can be obtained via physically realistic simulation of rendering of depth images.

In a geometric perspective, we argue that point distribution shift can be decomposed into 
1) shape changes of each point along the normal direction of surface due to all kinds of noises and 2) points' density changes along their tangent plane (\textit{i.e.}, approximately equivalent to the object surface).
Such a geometric interpretation of Sim2Real domain gap encourages us to propose an integrated scheme for domain-adapted feature learning -- {Quasi-balanced Self-training on Speckle-projected Synthesis (QS$^3$)}, which consists of physically realistic 3D synthesis of realistically systematic noises causing shape changes and feature encoding with domain-invariant patterns on density variations. 
Specifically, we introduce 1) a Mesh-to-Point data generation based on photorealistic rendering of stereo images with projection of speckle patterns, which generates a new synthetic point cloud dataset -- SpeckleNet 
(see Sec. \ref{subsec:M2P}), and 2) a novel self-training method on our SpeckleNet data to mitigate the suffering from unknown data distribution of target domain (see Sec. \ref{subsec:QBST}).

\subsection{Synthesis of Realistic Point Clouds via Physical Simulation 
}\label{subsec:M2P}

Previous works \cite{ICL-NUIM,kinectbarron,kinectbohg,nguyen2012modeling} have shown neither theoretical noisy models nor data augment strategies can cover a diversity of noises appearing in real-world depth sensor scanning, \textit{e.g.}, axial noise, shadow noise and structural noise, which are induced by scene illumination, object material, hardware of sensors and composition of the scanned scene (see the survey \cite{mallick2014characterizations} for more details). 
Different from existing UDA methods on point cloud classification only enforcing feature alignment across domains, we argue that the synthesis of object point clouds sharing similar statistical distribution with target data is a powerful alternative, via simulating realistic non-uniform noises, which can be viewed as a special alignment in the input space. 
Inspired by recent works \cite{blensor,planche2017depthsynth,planche2021physics} utilizing Physical Based Rendering (PBR) for depth sensor simulation, we leverage photorealistic rendering to reproduce realistic noises via a virtual depth sensor. 
As consuming RGB-D sensors (\textit{e.g.}, Microsoft Kinect V1, Intel RealSense) actively projecting speckle patterns are widely used for indoor scene scanning, we build a virtual active stereo based depth scanner based on PBR. 
Note that, physical simulation of the depth sensor is not limited to projection of speckle patterns adopted in our paper, which can be replaced by other styles of depth sensors such as time-of-flight depth sensors or structured light based depth sensors via fringe pattern projection. 

\begin{figure}[t]
  \centering
   \includegraphics[width= 1.0\linewidth]{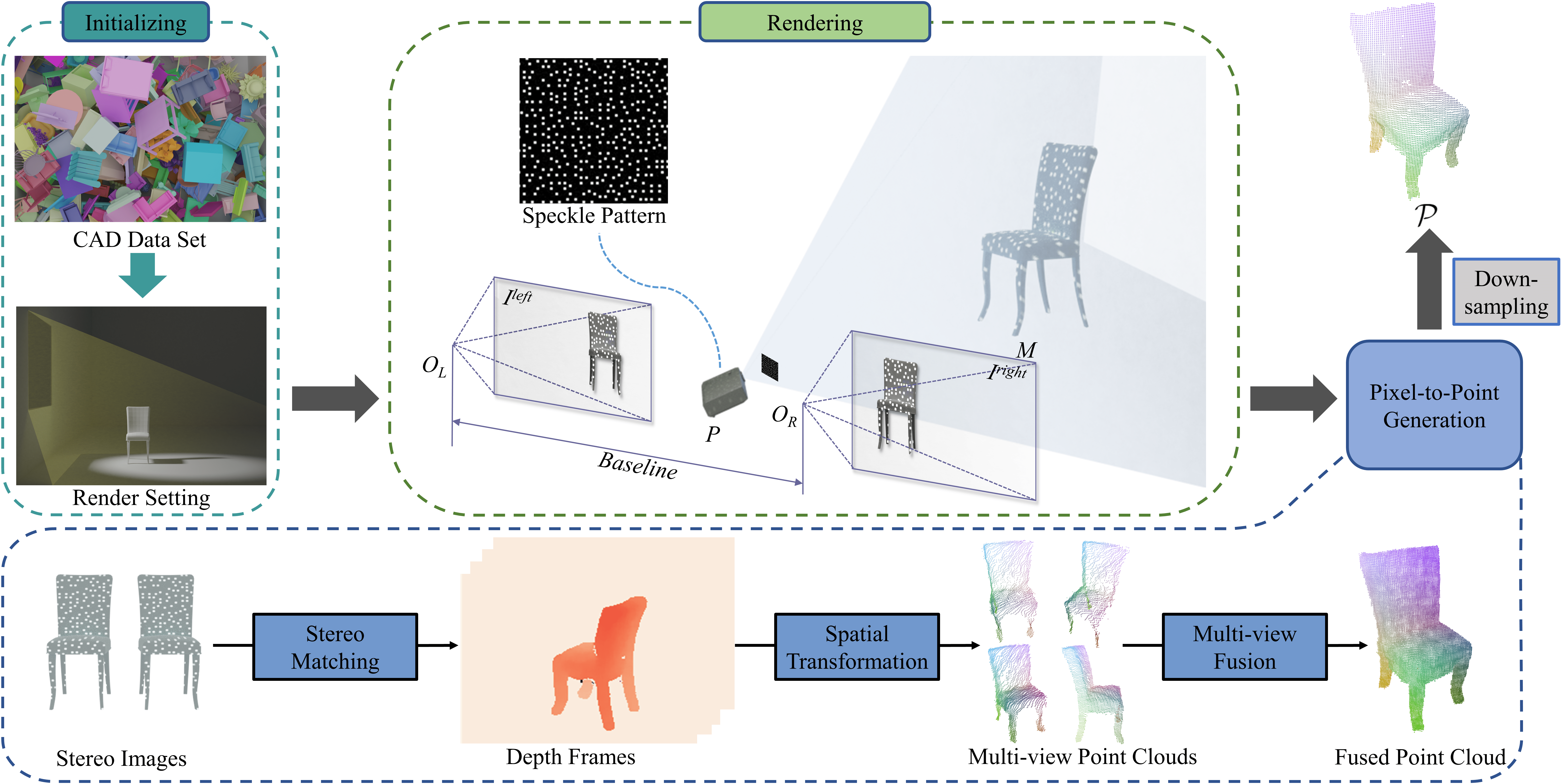}
   \caption{
   {Pipeline of the Mesh-to-Point method.}
During data pre-processing, we first scale the mesh-based model $\mathcal{M}$  to a unit-cube with an arbitrary rotation along the $z$-axis, which are fed into the Rendering module together with the settings of scene illumination and object reflection. 
In the Rendering module, we virtually organize a stereo camera and a projector. 
Given the model $M$, the projector $P$ actively projects a speckle pattern to $M$, then the stereo camera ($O_{L}$, $O_{R}$) outputs stereo images ${I}^\text{left}$ and ${I}^\text{right}$ to the next Pixel-to-Point Generation module. 
Depth images are converted via stereo matching on stereo images ${I}^\text{left}$ and ${I}^\text{right}$, which are further projected into 3D space using intrinsic parameters and extrinsic poses of the camera. 
Synthetic point clouds under multiple views are fused to produce a dense one carrying richer geometric information, which is then down-sampled to a sparse point set $\mathcal{P}$ as the final output.}
   \label{Pipeline}
\end{figure}

For replicating a typical active stereo based depth sensor, we organize a stereo camera and a projector, as an active illuminant to project pre-defined speckle pattern, in a simulation platform, \textit{e.g.}, Blender \cite{blender} adopted in this paper. 
The whole pipeline of the Mesh-to-Point synthesis is shown in Fig. \ref{Pipeline}.
In details, we set two identical optical imaging sensors (\textit{i.e.}, left camera $O_{L}$ and right camera $O_{R}$ in Fig. \ref{Pipeline}) for a stereo camera, where the image planes of two cameras are coplanar and translation between two cameras only along the $x$-axis of the left camera. 
The intrinsic and extrinsic parameters of the stereo camera are defined as prior knowledge in our virtual depth sensor. 
A projector $P$ is positioned in the middle of stereo cameras, actively projecting pre-defined speckle patterns on object surface to provide visual appearance, which thus benefits for discovering pixel correspondence in stereo matching. 
Given the CAD set $\mathcal{M}$ 
together with their labels $y \in \mathcal{Y}$ as input, the bidirectional scattering distribution function (BSDF) material \cite{bartell1981BSDFtheory} is adopted to model the scattered pattern of light by a surface, by following the default setting of BSDF function in \cite{blender} to initialize object models' material. We further place one area light source on top of the object model to be rendered, which can thus provide a uniform scene illumination. 
Note that, our paper concerns on simulating systematic noises due to modules' precision, with an empirically simplified rendering condition. 

With the aforementioned settings, pairs of RGB images $\{{I}_i^\text{left}, {I}_i^\text{right},y_i\}_{i=1}^{n_s}$ for each CAD model $M$ can be generated via photorealistic rendering. 
For generating a point cloud $\mathcal{P}$, we firstly perform stereo matching \cite{szeliski2010computervision} on the stereo RGB images ${I}^\text{left}$ and ${I}^\text{right}$ to gain disparity maps, each element $d$ of which can be used to compute its depth distance as follows \cite{szeliski2010computervision}: $(f \cdot b)/{d}$, where $f$ and $b$ are the focal length and the baseline of the camera. 
We further project depth images into 3D space to generate point clouds using intrinsic parameters and extrinsic poses of the camera. 
To mimic generation of realistic point clouds from RGB-D videos scanning object in multiple poses in practice, synthetic point clouds under multiple camera poses are fused together to produce dense sets, which are then down-sampled to the final point set $\mathcal{P}$.

\noindent \textbf{{SpeckleNet10} --} 
For a comparative evaluation, 
we use the same 10 categories of the ModelNet object models as PointDAN \cite{PointDAN}, and a new synthetic dataset, namely SpeckleNet10, can be generated via simulation of the above virtual active stereo based depth sensor.
Given the same size of CAD models as the ModelNet10, the only difference between ModelNet10 and our SpeckleNet10 lies in the procedure of the synthesis of object point clouds, \textit{i.e.}, reality of simulated noises. 
Parameter settings of our virtual depth sensor are empirically selected for typical real-world indoor scenarios instead of fitting a specific type depth sensor, based on the fact that practical noises of depth sensors are mainly affected by the types of projection patterns and the corresponding methods to compute depth, rather than extrinsic settings.
In our implementation, given an object model located in the scene, a stereo camera is randomly placed 3 to 5 meters away from the model, with an arbitrary elevation angle within $[20^\text{o}, 50^\text{o}]$ in the simulator, where the baseline distance $b$ between two imaging sensors in the stereo camera is set to 10 cm. For multi-view fusing under different viewing angles, one camera pose is randomly selected as an anchor to satisfy the aforementioned constraints, while the remaining as the variants of the anchored pose, lying in $(-10cm, 10cm)$ in translation and $(-0.1,0.1)$ in the Euler rotation.
With the setting of rendering, we can obtain stereo images with $1080\times 1080$ resolution {and utilize the block matching algorithm provided by the OpenCV to compute disparity from stereo images,} which are then converted to depth images with a down-sampling operation to $270*270$. 
As point clouds transformed from depth images remain rather dense,  sparse point sets of 2048 points are down-sampled via the Farthest Point Sample (FPS), as the input of UDA on point classification.

%-----------------------------------------
\subsection{Quasi-Balanced Self-Training}\label{subsec:QBST}

\begin{figure}[t]
	\centering
	\includegraphics[width=0.9\linewidth]{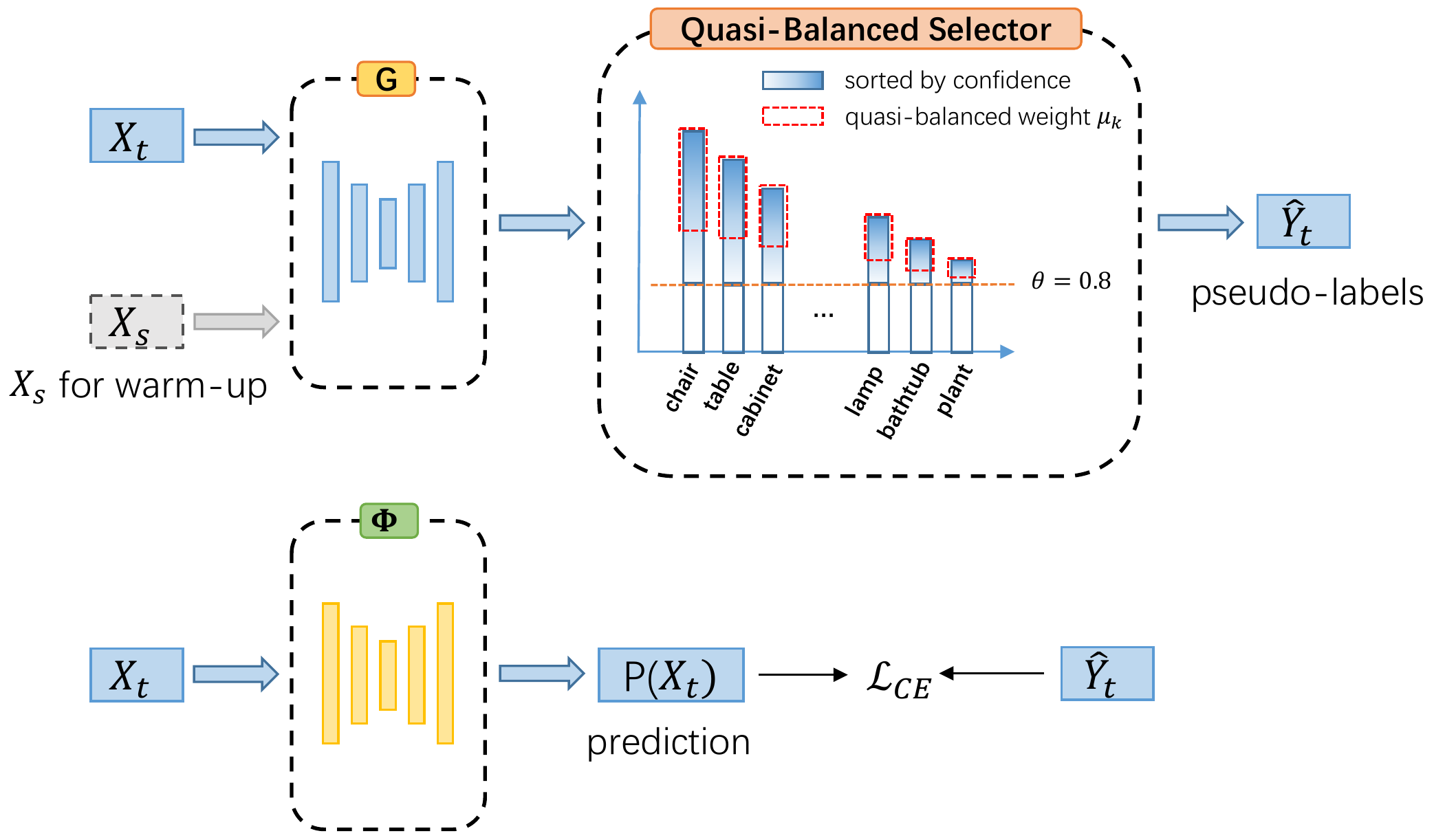}
	\caption{{Overview of our proposed \textit{Quasi-Balanced Self-Training}. \textbf{Top:} illustration of a data sparsity-driven strategy for pseudo label generation -- a quasi-balanced selector. 
	During self-training, labeled source data is only used for training a model as an initialized pseudo-label generator and does not participate in self-training. 
In the legend, the blue box indicates predicated confidence (the darker, the higher); the red dashed box denotes selection range by the quasi-balanced weight $\mu_k$, \textit{i.e.}, selecting a $\mu_k$ proportion of samples from the target samples classified as class $k$ with predicted confidence greater than $\theta$ (\textit{e.g.}, $\theta = 0.8$) as pseudo-label samples. \textbf{Bottom:} Training with selected pseudo-labels, where $\mathcal{L}_{CE}$ refers to the cross-entropy loss function.}}
	\label{Fig.QBST}
\end{figure}

Given the generated synthetic point clouds $X_s = \{\mathcal{P}^s\} \in \mathcal{S}$ with their corresponding class labels $Y_s = \{y^s\} \in \mathcal{Y}$ and real point clouds $X_t = \{\mathcal{P}^t\} \in \mathcal{T}$ respectively, the remaining part of domain-adapted feature learning shares the same setting as existing 3D UDA methods.
In this section, we propose a simple yet effective quasi-balanced self-training (QBST) method to dynamically select target instances for data balanceness when assigning pseudo labels in self-training.
An overview of our QBST is shown in Fig. \ref{Fig.QBST}, which consists of three steps below, with iteratively updating by the latter two.
\begin{itemize}
\setlength{\itemsep}{0pt}
\setlength{\parsep}{0pt}
\setlength{\parskip}{0pt}
\item [(a)] In the \textit{warm-up}, a supervised classification uses labeled source data (i.e., $\{X_s, Y_s\}$) to train a model \bm{$\Phi_o$} as an initialized pseudo-label generator \bm{$G_o$}.
\item [(b)] In the \textit{target instance selection with pseudo labels}, \bm{$G$} is used to obtain class prediction of unlabeled target data, and confident prediction will be assigned as pseudo labels for selected target samples.
\item [(c)] In the \textit{self-training}, instead of fine-tuning the {$\Phi_o$} as \cite{GAST}, an initial model {$\Phi_{init}$} is trained from scratch by selected target point clouds with pseudo-labels.
\end{itemize}

\noindent\textbf{Pseudo-label Generation --}
The generic pseudo-label generation strategy can be simplified to the following form when the model parameter $w$ is fixed:
\begin{equation*}\label{EqnQBSTLoss}
\begin{aligned}
&\min_{\widehat{Y}_t} ~~\mathcal{L}_{CE}(w, \widehat{Y}_t) = - \frac{1}{|X_t|} \sum_{\mathcal{P}_t \in X_t}  \sum_{k=1}^K \hat{y}_t^{(k)} \log {\frac{p(k|\mathcal{P}_t, w)}{\theta}} \\
&\quad {\rm s.t.}~~ \hat{y}_t \in \{[\text{onehot}]^K\} \cup {0}, \forall \hat{y}_t \in \widehat{Y}_t
\end{aligned}
\end{equation*}
where $\theta$ indicates the confidence threshold, and $\hat{y}_t = [\hat{y}_t^{(1)}, \hat{y}_t^{(2)},...,\hat{y}_t^{(K)}]$ is required to be a one-hot vector or all-zero vector. Therefore, $\hat{y}_t^{(K)}$ can be solved:

\begin{equation*}\label{EqnTrgLblAssign}
\hat{y}^k_t =\! \left\{
\begin{aligned}
& 1, \:\: {\rm if} ~\: k = \arg\max_{k} p(k|\mathcal{P}_t, w) \: ~\text{and}~ \ p(k|\mathcal{P}_t, w) > \theta \\
& 0, \:\: {\rm otherwise}.
\end{aligned}
\right.
\end{equation*}

Inspired by self-paced self-training (SPST) \cite{cbst}, we adopt a threshold $\theta$ that gradually increases with self-training iterations evolve (\textit{i.e.}, each iteration increases by a constant $\epsilon$, with more details in Alg. 1 of supplementary material).
Target instances to be assigned with pseudo-labels are selected by following a class-balanced rule \cite{cbst}: the more sparsity of data in a class $k$ is, the higher weight.
To this end, we generate target data distribution from the {pseudo-labeled samples selected according to data sparsity.}
In other words, unlabeled target samples that are predicted as class $k$ are sorted by confidence and selected in descending order of confidence according to the weight of
$\mu_k = (1 - \frac {L_k}{L})$,
where $L_k$ denotes the number of unlabeled target samples classified into class $k$, and $L$ is the total number of all unlabeled target samples whose predicted confidence greater than $\theta$.
{In contrast, CBST \cite{cbst} selects samples with the same proportion for all object classes, which cannot avoid label noises caused by low-confident mis-classified samples of long-tailed classes.
To avoid negative effects of imbalanced data distribution of source data, only selected target samples with pseudo labels are self-trained in our QBST, which is again different from the CBST.}

    \section{Experiments}

\subsection{Data and Settings}

\noindent \textbf{{Benchmarking Data in the Sim2Real UDA setting  --}}
{We construct a new and challenging benchmark for evaluating point cloud classification on Sim2Real domain adaptation, where a synthetic point cloud data is generated as source domain while a real-world one is set as target domain. 
Inspired by PointDA-10 \cite{PointDAN}, a pioneering Sim2Real benchmark, we directly use the CAD models of ModelNet10 from the PointDA-10 to generate synthetic data as source domain. 
Different from point clouds uniformly sampled from the CAD models of object classes as the ModelNet10, realistically-simulated point clouds generated by our Mesh-to-Point are collected as the SpeckleNet presented in Sec. \ref{subsec:M2P}.  
The PointDA-10 builds a real ScanNet10 as target domain via extracting the same 10 classes instances from ScanNet \cite{scannet}, where point clouds are directly generated from mesh vertices of the reconstructed surface. 
As meshing in shape reconstruction can alleviate realistic sensor noises, the point clouds in original ScanNet10 cannot reflect the true challenge of severe noises in practical classification of object point clouds. 
Moreover, meshing of partial and noisy point clouds is challenging and demands extra processing, which is typically unsuitable for real-time perception. 
To this end, we follow the setting in the PointDA-10 and choose the same scenes from the ScanNet to generate a more challenging real-world dataset -- DepthScanNet10, directly from the raw outputs of depth sensors. 
Specifically, to gain the point cloud of a single object instance, 2D instance segmentation is first applied to crop a depth image patch of the instance from the whole frame, which can then be converted to a point cloud. 
Multi-view fusion is employed to gather point clouds from several different viewpoints, followed by 
down-sampling to produce a fixed size of 2048 points. 
Note that, only those more than 1000 points will be kept for multi-view fusion whose size of viewpoints is set to 10. 
}

\noindent \textbf{Settings --}
On the Sim2Real benchmarking data, our proposed pipeline is compared with recent methods. 
ModelNet10 (\textbf{M}) and SpeckleNet10 (\textbf{S}) are employed as source domain, while we choose testing split of DepthScanNet10 (\textbf{D})) as target domain, following the setting of the PointDA-10.
Specifically, all object point clouds are normalized within a unit ball and down-sampled to 1024 points using the FPS algorithm. Random rotation along the z-axis and jittering as \cite{PointNet} is employed for data augmentation during training.
Moreover, the same data split of the ScanNet10 is used in the DepthScanNet10. 
 
\noindent \textbf{Comparative Methods --}
We evaluate several UDA classification methods, including
Point Domain Adaptation Network (\textbf{PointDAN}) \cite{PointDAN}, Deformation Reconstruction Network with Point Cloud Mixup (\textbf{DefRec+PCM}) \cite{DefRecPCM} and Geometry-Aware Self-Training (\textbf{GAST}) \cite{GAST}, by following the settings of the best performance in their paper, for simulation-to-reality domain adaptation.  

\noindent \textbf{Implementation Details --}
{All the networks are implemented based on PyTorch \cite{pytorch}. 
Beyond the SpeckleNet10 generated by our Mesh2Point simulator, 
we utilize DGCNN \cite{DGCNN} with Point Cloud Mixup (PCM) \cite{DefRecPCM} as our network baseline, followed by our proposed QBST as the self-training strategy to perform feature alignment on the SpeckleNet10. 
To implement recent UDA methods on point classification, except for PointDAN \cite{PointDAN} using official released source codes, other UDA methods are implemented based on DGCNN \cite{DGCNN} as the backbone network. 
We choose the ADAM \cite{kingma2014adam} as our optimizer with an initial learning rate of 0.001 and weight decay of 0.00005 and an epoch-wise cosine annealing learning rate scheduler for the UDA methods as \cite{GAST}. 
With the cross-entropy loss, we train all the methods for 200 epochs with a batch size 16 on an NVIDIA GTX-1080 Ti GPU. 
In each iteration of self-training, we adopt the ADAM with learning rate $1 \times 10^{-3}$ and batch size 32 for 10 epochs. 
The initial threshold $\theta_0$ is set to 0.8, and the constant $\epsilon$ is set to $5 \times 10^{-3}$. 
The mean accuracy and standard error of the mean (SEM) is reported on three trials of random seeds.}

\subsection{Results}
%-----------------------------------------------------

\begin{table}[t]
\caption{Effects of simulation of realistic noises with real data DepthScanNet (\textbf{D}) on classification accuracy ($\%$).} % \caption
\centering
\resizebox{0.48\linewidth}{!}
{ %< auto-adjusts font size to fill line
\setlength\tabcolsep{5pt}
\begin{tabular}{@{}lcc@{}}
\toprule
Method 
& M$\rightarrow$D  & S$\rightarrow$D\\
\midrule
Pointnet++ \cite{Pointnet++} 

& 48.4 $\pm$ 1.3  & \textbf{60.9 $\pm$ 0.8}  \\
DGCNN \cite{DGCNN} 
 
& 46.7 $\pm$ 1.4 & \textbf{64.0 $\pm$ 1.0} \\
RSCNN \cite{RSCNN} 

& 49.7 $\pm$ 1.1 & \textbf{53.9 $\pm$ 0.2} \\
SimpleView \cite{SimpleView} 
 
& 54.6 $\pm$ 0.7 & \textbf{62.3 $\pm$ 1.3} \\
\bottomrule
\end{tabular}
}
\label{tab:noise}
\end{table}

\noindent\textbf{Effects of the Synthesis of Realistic Noises with Ordinary Point Cloud Classifiers --}
We report comparative evaluation on real data \textbf{D} of ordinary point cloud classification with four popular classifiers in columns of M$\rightarrow$D and S$\rightarrow$D of Table \ref{tab:noise}.
Following recent SimpleView \cite{SimpleView},  the following four classification networks, \textit{i.e.}, PointNet++ \cite{Pointnet++}, DGCNN \cite{DGCNN}, RSCNN \cite{RSCNN} and SimpleView \cite{SimpleView}, are evaluated and compared.
It is evident that all the methods training on synthetic point clouds (from the \textbf{S}) of the Mesh2Point method can perform better than those on simply-augmented synthetic data (from the \textbf{M}), \textit{i.e.}, results in the column of \textbf{S$\rightarrow$D} are significantly superior to those in the column of  \textbf{M$\rightarrow$D} in Table \ref{tab:noise}, which demonstrate our motivation of using physical simulation to alleviate Sim2Real domain gap, owing to capturing realistic noises underlying in real data.

%-----------------------------------------------------

\begin{table}[t]
\caption{
Comparative evaluation in classification accuracy (\%) averaged over 3 seeds ($\pm$ SEM) on the Sim2Real data with recent UDA methods.
} % \caption
\centering
\resizebox{0.58\linewidth}{!}
{ %< auto-adjusts font size to fill line
\setlength\tabcolsep{5pt}
\begin{tabular}{@{}lccccc@{}}
\toprule
Method & M$\rightarrow$D & S$\rightarrow$D \\
\midrule
Supervised  & 90.4 $\pm$ 0.4 & 90.4 $\pm$ 0.4 \\
DGCNN \cite{DGCNN} (w/o Adapt)  & 46.7 $\pm$ 1.4 & {64.0 $\pm$ 1.0} \\
\midrule
PointDAN \cite{PointDAN}  & 58.9 $\pm$ 0.9 & {62.9 $\pm$ 1.6} \\
DefRec \cite{DefRecPCM}  & 57.8 $\pm$ 1.1 & {60.8 $\pm$ 0.6} \\ 
DefRec+PCM \cite{DefRecPCM}  & 62.1 $\pm$ 0.8 & {64.4 $\pm$ 0.7} \\
GAST w/o SPST \cite{GAST}  & {62.4 $\pm$ 1.1} & 61.8 $\pm$ 1.0 \\
GAST \cite{GAST} & {64.8 $\pm$ 1.4} & 64.4 $\pm$ 0.2 \\
QBST (ours) & \textbf{66.4} $\pm$ 1.1 & -- \\
QS$^3$ (ours) & -- & \textbf{72.4} $\pm$ 0.8\\
\bottomrule
\end{tabular}
} % \resizebox

\label{tab:uda-core}
\end{table}

\noindent\textbf{Rationale of Balanced Data Distribution for UDA Point Cloud Classification --}
In Table \ref{tab:uda-core}, a number of recent UDA methods are evaluated and compared on simulation-to-reality tasks.
The \textbf{Supervised} method trains DGCNN \cite{DGCNN} backbone with labeled target data only, and the \textbf{DGCNN w/o Adapt} method trains the identical net with only labeled source samples, treated as a reference of the upper and lower performance bound respectively.
On one hand, models training on the physically simulated instances (from the \textbf{S}) cannot be consistently superior to those on augmented ones (from the \textbf{M}).
Specifically, superior performance of DefRec, GAST and its degenerated GAST w/o SPST with the \textbf{M} to the proposed \textbf{S} when evaluation on real data \textbf{D}, where domain gap of \textbf{M}$\rightarrow$\textbf{D} is verified to be larger than those of \textbf{S}$\rightarrow$\textbf{D} as shown in Table \ref{tab:noise}. 
Such a phenomena can be explained by that these self-supervised methods (\textit{i.e.,} DefRec and GAST) rely on capturing domain-invariant geometric patterns from local regions, where synthetic data in the \textbf{S} are generated from partially visible surfaces and thus suffers more than those complete ones in the \textbf{M} to learn cross-domain local geometric patterns.
DefRec+PCM can significantly improve performance on \textbf{S}$\rightarrow$\textbf{D} over DefRec, which can be credited to the extra PCM data augmentation to enrich data diversity and inherently balance data distribution by increasing samples shared with long-tailed classes.  
On the other hand, it is observed that all the UDA methods on  \textbf{M}$\rightarrow$\textbf{D} can outperform the backbone DGCNN, while very marginal improvement or even negative transfer performance of most of methods can be gained on the remaining task.
With the PCM and SPST modules respectively, negative transfer on DefRec and GAST can be alleviated, which can confirm their effectiveness.
As \cite{GAST}, the SPST cannot guarantee the correctness of pseudo labels assigned to target instances but under the assumption that they are mostly correct, which are largely affected by performance of instance selector.
Inspired by the success of PCM , it is encouraged to propose a self-training method to achieve balanced data distribution. 
%-----------------------------------------------------

\noindent\textbf{Effects of Quasi-balanced Self-Training --} 
With the DGCNN as the identical backbone, 
we compare our QBST with its direct competitors -- {SPST and CBST} as well as PCM, whose results are reported in Table \ref{tab:st-core}.
As aforementioned, the PCM and SPST can effectively inhibit negative transfer, it is observed that {both SPST and CBST} can work better together with the PCM owing to improvement on selection of confident target samples gained by the PCM.
In light of this, based on the backbone DGCNN and the PCM, superior performance of the proposed QBST can confirm its effectiveness by consistently beating its direct competitor SPST and CBST as well as the state-of-the-art DefRec+PCM and GAST (refer to Table \ref{tab:uda-core}) for both Sim2Real tasks.
Superior performance of our QBST to SPST and CBST can be credited to using data sparsity-sensitive weight sampling on high-confident target samples, while comparative methods (i.e., SPST and CBST) cannot alleviate negative effects from mis-classified samples of long-tailed classes with low confidence.

\begin{table}[t]
\caption{
Effects of our QBST with classification accuracy (\%) averaged over 3 seeds ($\pm$ SEM) on the sim2real data.
} % \caption
\centering
\resizebox{0.58\linewidth}{!}
{ %< auto-adjusts font size to fill line
\setlength\tabcolsep{5pt}
\begin{tabular}{@{}lccc@{}}
\toprule
Method 
& M$\rightarrow$D & S$\rightarrow$D \\
\midrule
Supervised 
& 90.4 $\pm$ 0.4 & 90.4 $\pm$ 0.4 \\
DGCNN \cite{DGCNN} (w/o Adapt)  
& 46.7 $\pm$ 1.4 & 64.0 $\pm$ 1.0 \\
\midrule
DGCNN+CBST
& 41.8 $\pm$ 2.0 & 58.6 $\pm$ 1.7 \\ 
DGCNN+SPST 
& 45.2 $\pm$ 0.5 & 63.3 $\pm$ 0.5 \\
DGCNN+QBST
& 45.6 $\pm$ 0.4 & 63.8 $\pm$ 0.1 \\
\midrule
DGCNN+PCM 
& 61.2 $\pm$ 0.6 & 68.5 $\pm$ 1.1 \\
DGCNN+PCM+CBST
& 62.5 $\pm$ 3.0 & 62.9 $\pm$ 0.9 \\
DGCNN+PCM+SPST 
& 65.4 $\pm$ 0.6 & 71.9 $\pm$ 0.7 \\
DGCNN+PCM+QBST 
& \textbf{66.4 $\pm$ 1.1} & \textbf{72.4 $\pm$ 0.8}\\
\bottomrule
\end{tabular}
} % \resizebox

\label{tab:st-core}
\end{table}

\noindent \textbf{{Comparison with the State-of-the-art Methods --}} Our QS$^3$ scheme and the degenerated QBST are compared with recent UDA methods, whose results are shown in Table \ref{tab:uda-core}. 
It is evident that classification accuracy of our QS$^3$ scheme can reach 72.4\%, a large marginal improvement over other UDA methods, when testing on the DepthScanNet.
Using our QBST on the ideal synthetic point clouds from the ModelNet (\textit{i.e.,} M$\rightarrow$D), superior performance to the state-of-the-art UDA methods can still be achieved, which can again verify the effectiveness of the proposed QBST.

%---------------------------------------------------------------------------

\begin{table}[t]
\caption{Ablation studies about simulation of realistic noises of stereo rendering and matching with real dataset \textbf{D} on classification accuracy ($\%$).} % \caption
\centering
\resizebox{0.75\linewidth}{!}
{ %< auto-adjusts font size to fill line
\setlength\tabcolsep{5pt}
\begin{tabular}{@{}lcccc@{}}
\toprule
Method 
& S$_c$$\rightarrow$D & B$\rightarrow$D  &  M$_d$$\rightarrow$D & S$\rightarrow$D\\
\midrule
Pointnet++ \cite{Pointnet++} 

 & 50.2 $\pm$ 2.0 & 52.4 $\pm$ 1.3 & 57.9 $\pm$ 1.2 & \textbf{60.9 $\pm$ 0.8}  \\
DGCNN \cite{DGCNN} 
 
 & 53.2 $\pm$ 1.6 & 56.6 $\pm$ 2.1 & 50.4 $\pm$ 1.0 & \textbf{64.0 $\pm$ 1.0} \\
RSCNN \cite{RSCNN} 

 & 48.8 $\pm$ 0.1 & 51.7 $\pm$ 2.0 & \textbf{56.5 $\pm$ 1.0} & 53.9 $\pm$ 0.2 \\
SimpleView \cite{SimpleView} 
 
 & 56.2 $\pm$ 1.2 & \textbf{65.1 $\pm$ 0.2} & 57.4 $\pm$ 0.7 & 62.3 $\pm$ 1.3 \\
\bottomrule
\end{tabular}
}
\label{tab:noise_ablation}
\end{table}

\noindent\textbf{Ablation Studies about Simulation of Realistic Data with Ordinary Point Cloud Classifiers 
--} 
Motivation of our physically rendering is to simulate non-neglected systematic noises, which is considered as the key factor to reduction of shape shift in Sim2Real domain gap.
To verify it, we generate a clean depth image, whose pixels' depth value are directly measured according to distance between the corresponding point on object surface and the camera along its optical axis, to avoid depth computation via stereo imaging and matching. 
Synthetic data from these ideally clean depth frames are termed as a SpeckleNet10-Clean (\textbf{S}$_c$). 
Some works \cite{planche2017depthsynth,planche2021physics,blender} on sensor simulation can be adopted in our scheme to replace the Mesh2Point method.
Therefore, we compare one of the representative simulators -- Blensor \cite{blensor}, an open-source depth sensor simulator based on Blender \cite{blender}, which output a set of point clouds (namely Blensor10, \textbf{B}) using the same setting as our Specklenet10. 
Moreover, we also provide a better baseline for the ModelNet10 where we perform random region dropout on point clouds to simulate missing parts typically encountered in real data as further data augmentation. Such an augmented dataset is named as ModelNet10-Dropout (\textbf{M}$_d$). 
Results in Table \ref{tab:noise_ablation} can demonstrate superior performance of most of models training on the SpeckleNet10 (\textbf{S}), which can be credited to simulation of realistic noises. 
Good performance of the RSCNN on  M$_d\rightarrow$D and the SimpleView on B$\rightarrow$D can verify our claim that density shift should not be omitted beyond systemic noises of the depth sensor.

    %%%%%%%conclusion
\section{Conclusion}
This paper investigates an effective pipeline to mitigate Sim2Real domain gap in a novel perspective of physically realistic synthesis of object point clouds. 
The synthesis of realistic data can inherently be robust against imbalanced data distribution,
as we can observe negative transfer of existing UDA methods in our experiments, which encourages us to propose the quasi-balanced self-training for further suppression.
Experiment results can verify the effectiveness of the unified QS$^3$ scheme as well as both physical simulation of realistic noises and also our quasi-balanced self-training, achieving the state-of-the-art performance on the challenging Sim2Real domain adaptation tasks.
Moreover, the proposed Mesh2Point method is not differentiable and parameter settings of the synthesis pipeline are empirically selected, which could cause sub-optimal generation for specific tasks. 
Inspired by DDS \cite{planche2021physics}, incorporating differentiable physics-based rendering into our QS$^3$ scheme in an end-to-end learning manner can be a promising direction in future.
    \subsubsection{Acknowledgements}
This work is supported in part by the National Natural Science Foundation of China (Grant No.: 61902131), the Program for Guangdong Introducing Innovative and Enterpreneurial Teams (Grant No.: 2017ZT07X183), the Guangdong Provincial Key Laboratory of  Human Digital Twin (Grant No.: 2022B1212010004). 
    \bibliographystyle{splncs04}
    \bibliography{main}
\end{document}